%% file: 0.example_paper.tex
\documentclass{article}

\usepackage{microtype}
\usepackage{graphicx}
\usepackage{subcaption}
\usepackage{booktabs} 

\usepackage{hyperref}

\usepackage[accepted]{icml2026}

\usepackage{amsmath}
\usepackage{amssymb}
\usepackage{mathtools}
\usepackage{amsthm}

\usepackage{amssymb}
\usepackage{booktabs}
\usepackage{subcaption}
\usepackage{multirow}
\usepackage{wrapfig} 
\usepackage{colortbl}
\usepackage{enumitem}
\usepackage{kotex} 
\usepackage[most]{tcolorbox}
\usepackage{adjustbox}

\usepackage[capitalize,noabbrev]{cleveref}

\theoremstyle{plain}

\theoremstyle{definition}

\theoremstyle{remark}

\usepackage[textsize=tiny]{todonotes}

\icmltitlerunning{Manga109-v2026: Revisiting Manga109 Annotations for Modern Manga Understanding}

\begin{document}

\twocolumn[
  \icmltitle{Manga109-v2026: Revisiting Manga109 Annotations \\ for Modern Manga Understanding}
  
    \icmlsetsymbol{equal}{*}
    
    \begin{icmlauthorlist}
        \icmlauthor{Jeonghun Baek}{equal,utokyo}
        \icmlauthor{Atsuyuki Miyai}{equal,utokyo}
        \icmlauthor{Shota Onohara}{equal,utokyo}
        \icmlauthor{Hikaru Ikuta}{mantra}
        \icmlauthor{Kiyoharu Aizawa}{utokyo}
    \end{icmlauthorlist}
    
    \icmlaffiliation{utokyo}{
    The University of Tokyo,
    Tokyo, Japan
    }
    
    \icmlaffiliation{mantra}{
    Mantra Inc., Tokyo, Japan
    }
    
    \icmlcorrespondingauthor{Jeonghun Baek}{baek@hal.t.u-tokyo.ac.jp}

    \vspace{0.1cm}
    \begin{center}
    Project page: \url{https://manga109.github.io/manga109-project-website/en/}
    \end{center}
    
  \icmlkeywords{Machine Learning, ICML}

  \vskip 0.3in
]



\printAffiliationsAndNotice{\icmlEqualContribution}

\input{1.body}

\section*{Acknowledgments}
In this research work, we used the UTokyo Azure~\cite{MakotoNakamura20250030}.

\bibliography{example_paper}
\bibliographystyle{icml2026}

\end{document}

%% file: 1.body.tex
\begin{abstract}
Manga is a culturally distinctive multimodal medium and one of the most influential forms of Japanese popular culture. As AI systems increasingly target manga understanding, OCR, and translation, Manga109 has become a foundational dataset for manga-related AI research. However, the current Manga109 dataset contains inaccurate transcriptions and coarse annotations, which do not align well with modern OCR and multimodal manga understanding tasks. In this work, we revisit the dialogue text annotations of Manga109 and identify five categories of annotation issues, including inaccurate transcriptions, missing text regions, overlapping dialogue and onomatopoeia, and under-segmented speech balloons. To address these issues, we combine OCR-based issue detection and manual revision to construct Manga109-v2026, revising approximately 29,000 dialogue annotations. Our revisions better align Manga109 with modern OCR and multimodal manga understanding systems while preserving expressive structures characteristic of manga.
\end{abstract}

\begin{figure*}[t]
    \centering
    \includegraphics[width=\linewidth]{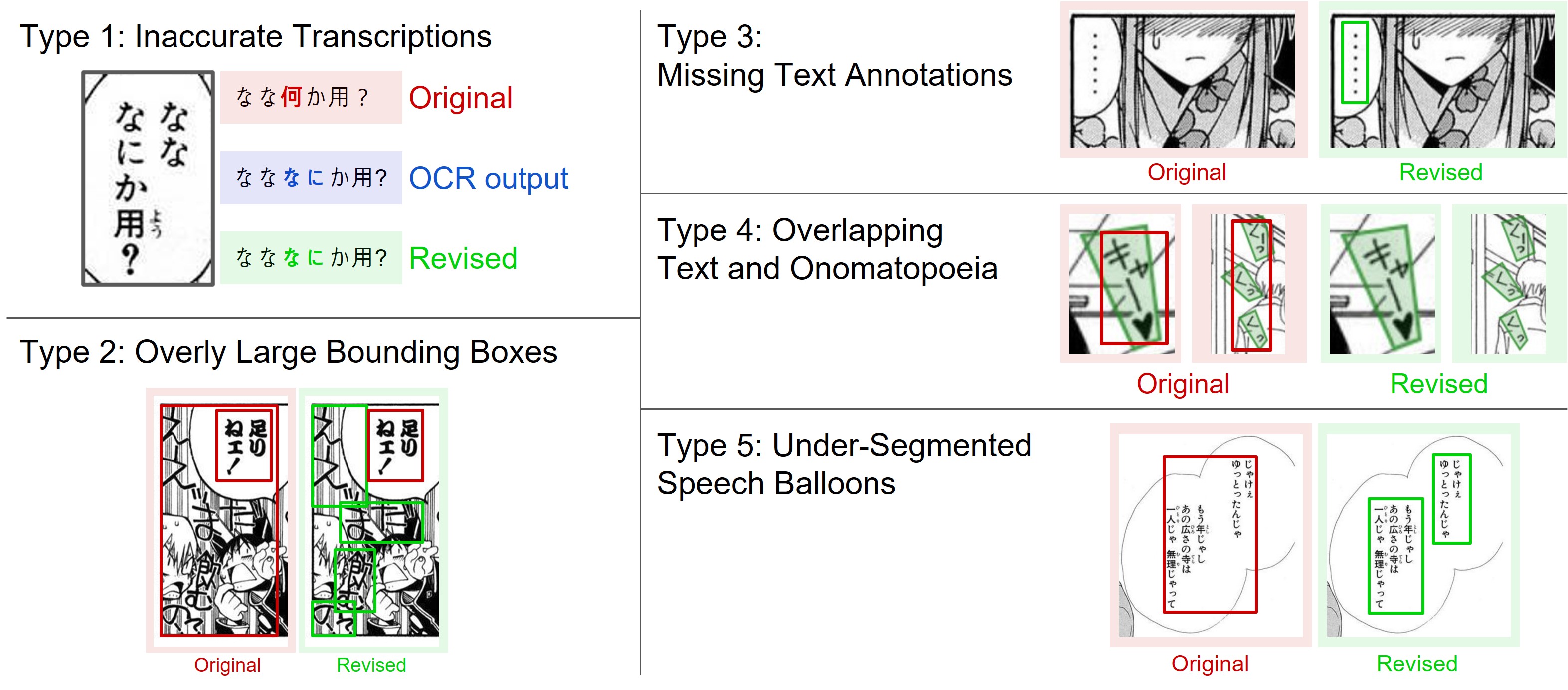}
    \vspace{-5mm}
    \caption{Overview of the five annotation issue types addressed in Manga109-v2026, together with representative examples of the original and revised annotations. 
    Images courtesy of Yamada Uduki, Inohara Daisuke, Akamatsu Ken, and Hasegawa Yuichi.}
    \label{fig:overview}
\end{figure*}

\section{Introduction}
Manga is a culturally distinctive multimodal medium that combines visual storytelling, expressive typography, speech balloon layouts, and stylized onomatopoeia. 
As one of the most influential forms of Japanese popular culture, manga has attracted increasing attention in AI research~\cite{vivoli2024one}, including optical character recognition (OCR), translation, multimodal understanding, and creator-support systems.

Manga109~\cite{manga109,manga109_2} is one of the most widely used datasets for manga-related AI research. 
Its rich annotations, including panel structures, character information, and dialogue text, have supported a broad range of downstream tasks such as OCR~\cite{aramaki2016text, baek2022COO,baek2025mangavqa}, translation~\cite{hinami2021towards}, transcript generation~\cite{magiv1, magiv2}, multimodal understanding~\cite{mangaub2025, baek2025mangavqa}, and manga-oriented large multimodal models~\cite{baek2025mangavqa}. 
As manga-oriented AI systems continue to advance, annotation quality increasingly affects not only model performance but also how AI systems interpret expressive structures characteristic of manga.

However, some annotation practices in Manga109~\cite{manga109_2}, while suitable for earlier OCR and manga analysis settings, can lead to mismatches with modern AI systems and downstream tasks.
For example, overlapping dialogue and onomatopoeia annotations may obscure stylistically meaningful sound expressions, while under-segmented speech balloons may conflict with how modern OCR systems interpret text regions. 
In addition, inaccurate transcriptions and missing annotations can reduce the reliability of evaluation and training data.
Such annotation issues may also affect future manga translation systems, where inaccurate or incomplete text annotations can lead to degraded translation quality and broader downstream issues.

In this work, we revisit the dialogue text annotations of Manga109 and identify five categories of annotation issues. 
To address these issues, we combine OCR-based issue detection and manual revision to construct Manga109-v2026, revising approximately 29,000 dialogue annotations, corresponding to 19.6\% of all text annotations in Manga109.
Our revisions better align Manga109 with modern OCR and multimodal manga understanding tasks. 
Experimental results show that the revised annotations substantially improve OCR evaluation performance, increasing evaluation scores by 14.4 percentage points.

More broadly, our work illustrates how AI and humans can collaboratively improve culturally grounded datasets by combining modern AI technologies with human verification. 
We hope these improvements support future AI-assisted OCR and translation systems that make manga more accessible across languages and cultural contexts.

\section{Related Work}
Manga has recently attracted increasing attention in multimodal AI research due to its unique combination of visual storytelling, stylized text, and expressive narrative structures. 
Manga109~\cite{manga109,manga109_2} has become a foundational dataset for manga-related AI tasks and has supported a wide range of downstream applications.

Several studies have extended Manga109 for specific tasks. 
MangaOCR~\cite{baek2025mangavqa} focuses on manga text recognition, while OpenMantra~\cite{hinami2021towards} targets manga translation. 
Other works address transcript generation~\cite{magiv1,magiv2}, multimodal manga understanding~\cite{mangaub2025, baek2025mangavqa}, speaker detection~\cite{li2024manga109dialog}, segmentation~\cite{MangaSeg}, and prose generation from manga images~\cite{magiv3}. 
Recent manga-oriented large multimodal models further explore the use of AI systems for understanding manga narratives and supporting manga-related creative applications~\cite{baek2025mangavqa}.

These studies demonstrate that Manga109 has become an important infrastructure for multimodal manga AI research. 
As downstream applications increasingly rely on structural and semantic understanding of manga, the quality and practices of annotation schemes become increasingly important.
In particular, expressive elements such as onomatopoeia, speech balloon layouts, and visually stylized text can significantly affect how AI systems interpret manga content.

In contrast to prior work that primarily builds downstream tasks and models on top of Manga109, our work revisits the annotation layer itself. 
We focus on improving dialogue text annotations to better align Manga109 with modern OCR and multimodal manga understanding tasks while preserving expressive structures characteristic of manga.

\begin{figure*}[t]
    \centering
    \includegraphics[width=0.95\linewidth]{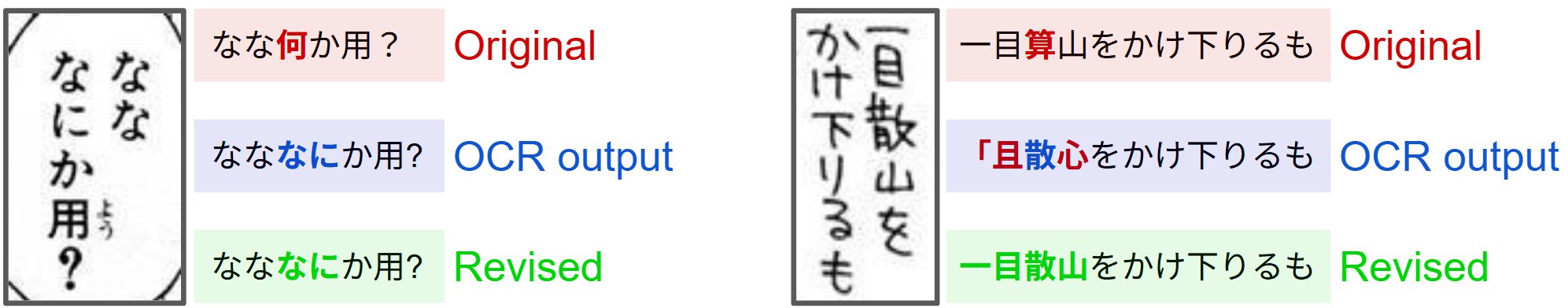}
    \caption{\textbf{Type 1: Inaccurate Transcriptions}, where the annotated text is incorrect.  
``Original'', ``OCR Output'', and ``Revised'' denote the original annotation, OCR output, and revised annotation, respectively. 
Red, blue, and green text indicate incorrect characters, correctly recognized OCR outputs, and revised characters, respectively. 
Images courtesy of Yamada Uduki.}
    \label{fig:type1}
\end{figure*}

\section{Annotation Issue Detection and Revision}
\subsection{Issue Detection Strategy}\label{sec:issue_detection}
We identify potential annotation issues by comparing existing Manga109 annotations with OCR outputs generated by a recent commercial API developed by Mantra Inc. for manga translation applications.
The OCR system itself is proprietary, and detailed architectural information is not publicly available.
According to Mantra Inc., the system was developed for practical manga translation applications with reference to prior manga OCR and translation research~\cite{hinami2021towards}. 
However, we found that its outputs often better reflected how modern AI systems interpret manga text regions and expressive structures.

Importantly, we do not treat OCR outputs as ground truth.
Instead, we manually inspect discrepancies between OCR outputs and existing annotations to identify cases where the legacy annotations may negatively affect modern downstream applications and evaluation settings. 
Through this process, we categorize the identified issues into five types and apply corresponding revisions.

\begin{table}[t]
    \tabcolsep=0.13cm
    \centering
    \caption{Number of annotation issues by type.}
    \label{tab:count}
    \begin{adjustbox}{width=\linewidth}
    \begin{tabular}[t]{@{}clr@{}}
        \toprule
        \textbf{Type} & \textbf{Name} & \textbf{Count} \\
        \midrule
        1 & Inaccurate Transcriptions & $\approx$ 9,200 \\
        2 & Overly Large Bounding Boxes & $\approx$ 50 \\
        3 & Missing Text Annotations & $\approx$ 800 \\
        4 & Overlapping Text and Onomatopoeia & $\approx$ 4,300 \\
        5 & Under-Segmented Speech Balloons & $\approx$ 14,900 \\
        \bottomrule
    \end{tabular}
    \end{adjustbox}
\end{table}

\subsection{Annotation Issue Types and Revisions}
\paragraph{Summary of Issue Types}
We categorize the identified annotation issues into five types: Type 1 (Inaccurate Transcriptions), Type 2 (Overly Large Bounding Boxes), Type 3 (Missing Text Annotations), Type 4 (Overlapping Text and Onomatopoeia), and Type 5 (Under-Segmented Speech Balloons).
Figure~\ref{fig:overview} provides an overview of these issue types, and Table~\ref{tab:count} summarizes the number of instances in each category. 
In total, we revised approximately 29,000 text annotations, corresponding to about 19.6\% of the 147,887 text annotations in Manga109.
The revision process involved manual verification and revision by four of the authors.

The five issue types can be grouped into three main objectives:
\begin{enumerate}[label=(\arabic*)]
    \item \textbf{Correcting annotation inconsistencies} introduced during the original dataset construction, corresponding to Types 1--3.
    
    \item \textbf{Improving compatibility with modern manga AI applications}, particularly in cases where overlapping or conflicting annotations affect the interpretation of expressive manga elements, corresponding to Type 4.

    \item \textbf{Revisiting legacy annotation practices} that were acceptable under earlier settings but have become problematic for modern downstream tasks and evaluation settings, corresponding to Type 5.
\end{enumerate}

Detailed descriptions of each type and the corresponding revisions are provided in the following paragraphs.

\paragraph{Type 1: Inaccurate Transcriptions}
This type refers to cases where the existing text annotations in Manga109 differ from OCR outputs (see Figure~\ref{fig:type1}).
In cases where OCR outputs and existing annotations disagree, determining the appropriate transcription may still require additional verification. 

To assist this process, we use two large language models, GPT-5~\cite{singh2025openai} and Gemini 3 Flash~\cite{google_gemini3flash_2025}. 
Given the original annotation and OCR output, the models are asked to select the more appropriate transcription. 
When both models select the same transcription, we treat it as the correct transcription, which occurred in 15,359 cases. 
Among these, 7,156 cases resulted in revisions where the OCR output was judged to be more appropriate than the original annotation, while the remaining 8,203 cases retained the original annotation. 
The remaining 2,051 disagreement cases were manually inspected and verified by the authors, resulting in a total of 9,207 revised annotations.

Correcting these inconsistencies improves the reliability of manga text annotations for downstream tasks such as OCR and translation.

\begin{figure}[t]
    \centering
    \includegraphics[width=0.95\linewidth]{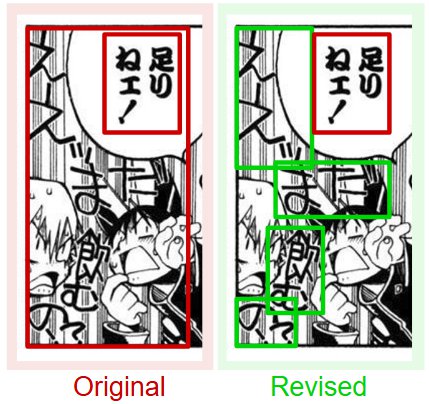}
    \caption{\textbf{Type 2: Overly Large Bounding Boxes}, where a large bounding box encloses other text regions or unrelated visual content.
Red and green bounding boxes denote the original and revised annotations, respectively. 
Images courtesy of Inohara Daisuke.}
    \label{fig:type2}
\end{figure}

\paragraph{Type 2: Overly Large Bounding Boxes}
This type refers to cases where a single text bounding box covers an excessively large region, often including other text instances or non-text visual elements within a speech balloon (see Figure~\ref{fig:type2}).

To identify such cases, we detect bounding boxes that overlap with other text regions. 
The resulting candidates are manually inspected, and overly large bounding boxes are divided into smaller regions corresponding to individual text instances while minimizing the inclusion of unrelated visual content.

This revision improves the spatial precision of text annotations, which is important for downstream tasks such as text localization and manga text understanding.

\begin{figure}[t]
    \centering
    \includegraphics[width=0.95\linewidth]{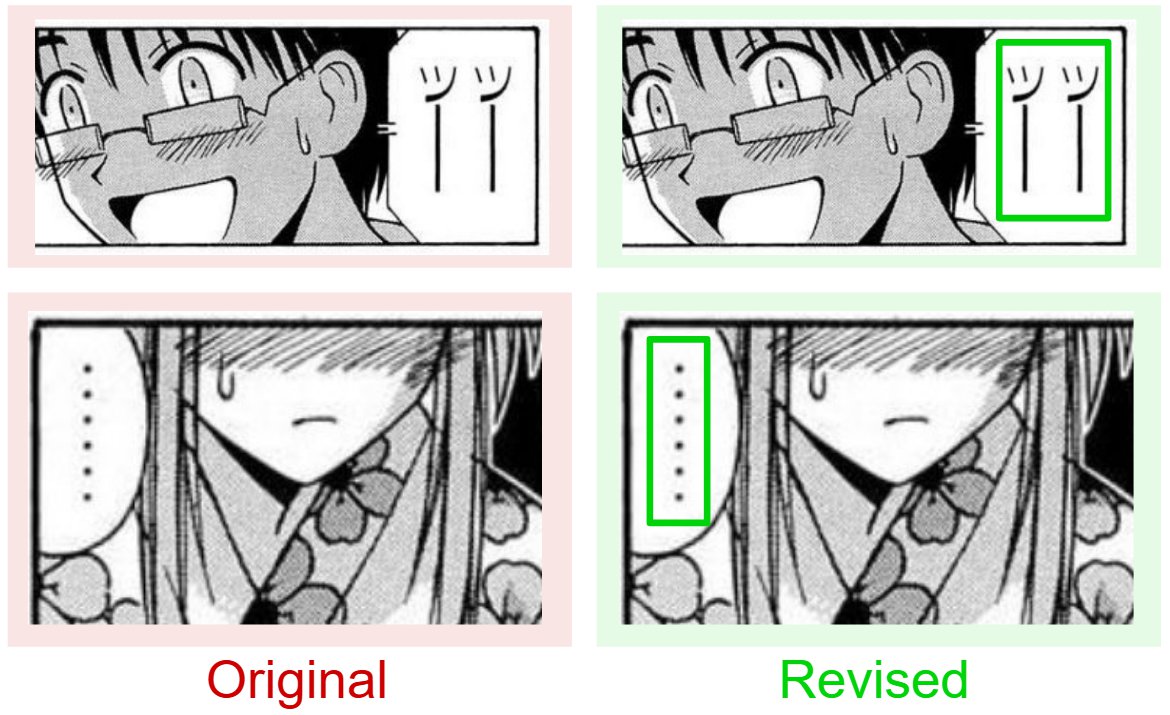}
    \caption{\textbf{Type 3: Missing Text Annotations}, where text regions are not annotated. 
    Images courtesy of Akamatsu Ken.}
    \label{fig:type3}
\end{figure}

\paragraph{Type 3: Missing Text Annotations}
This type refers to cases where text regions present in the image are not annotated in Manga109 (see Figure~\ref{fig:type3}).
We identify such cases by comparing OCR outputs with the existing annotations. 
In many cases, short expressions such as ``!'' or ``…'' are omitted. 
Although visually small, these expressions can still contribute to the pacing and expressive structure of manga narratives. 

We manually verify these cases and add the missing bounding boxes and corresponding transcriptions. 
This revision improves annotation completeness, which is important for reliable evaluation in downstream tasks such as OCR, translation, and manga understanding.

\begin{figure*}[t]
    \centering
    \includegraphics[width=0.9\linewidth]{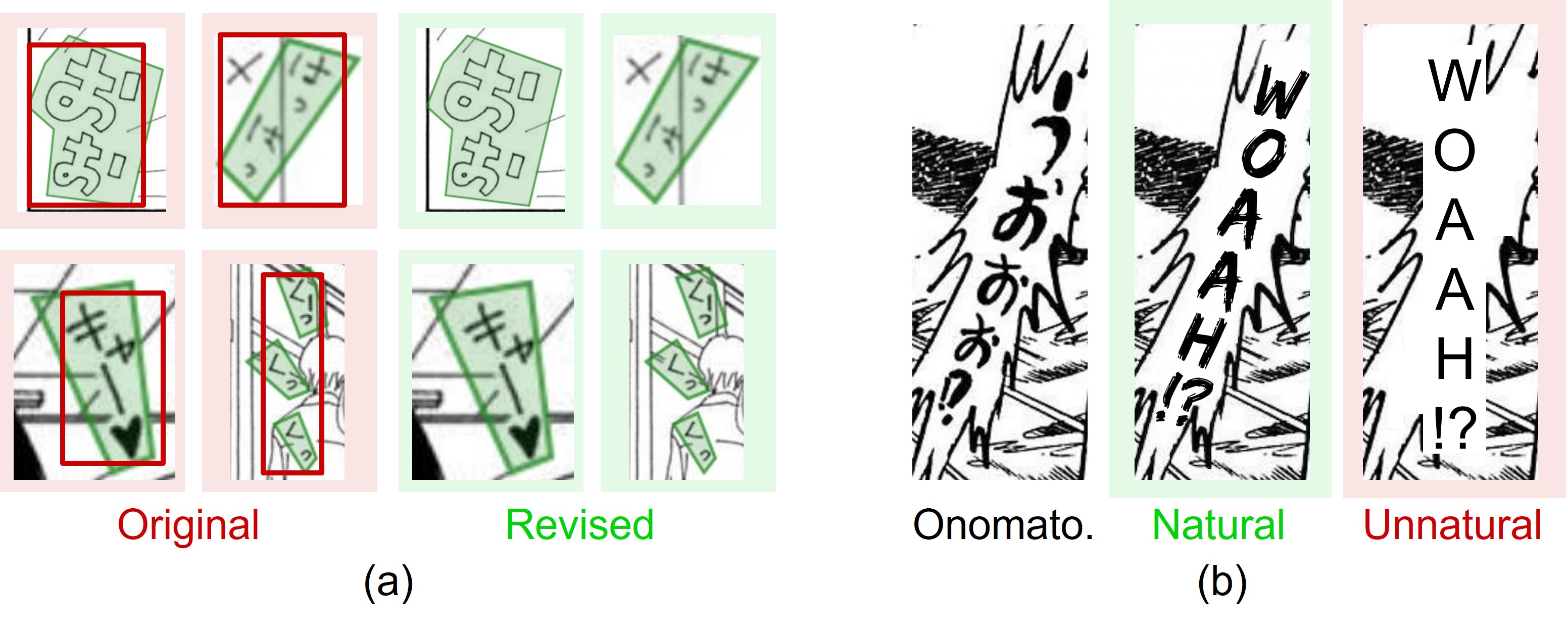}
    \vspace{-2mm}
    \caption{\textbf{(a) Type 4: Overlapping Text and Onomatopoeia Annotations}, where dialogue and onomatopoeia overlap. 
    Green polygon regions indicate onomatopoeia text.
    (b) ``Natural'' denotes translations that preserve the expressive and stylistic characteristics of onomatopoeia, while ``Unnatural'' denotes translations that treat onomatopoeia as regular dialogue without preserving their stylistic characteristics.
    Images courtesy of Yamada Uduki and Hasegawa Yuichi.}
    \label{fig:type4}
\end{figure*}

\paragraph{Type 4: Overlapping Text and Onomatopoeia Annotations}
Onomatopoeia annotations (i.e., annotations of onomatopoeic text in images) were introduced in 2022~\cite{baek2022COO}. 
However, in some cases, these annotations overlap with dialogue text regions, leading to inconsistencies in how expressive manga text is represented (see Figure~\ref{fig:type4}(a)).

In such cases, onomatopoeic text may be incorrectly treated as regular dialogue text, despite often serving stylistic and narrative functions distinct from ordinary dialogue. 
This can negatively affect downstream tasks such as translation. 
For example, treating onomatopoeia as regular dialogue without preserving its stylistic characteristics may result in unnatural translations, as illustrated in Figure~\ref{fig:type4}(b).

To address this issue, we identify overlapping regions based on their spatial positions. 
We retain onomatopoeia annotations and remove overlapping regions from dialogue text bounding boxes. 
This revision improves the consistency of expressive text annotations and better supports downstream tasks such as translation and manga understanding.

\begin{figure*}[t]
    \centering
    \includegraphics[width=0.9\linewidth]{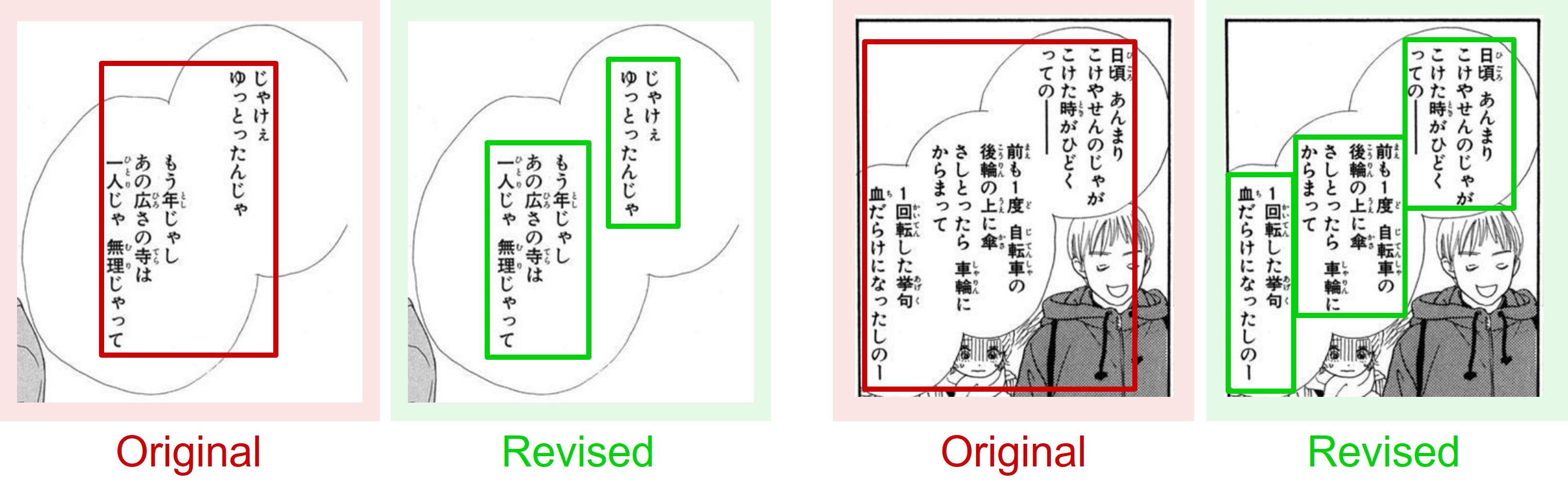}
    \caption{\textbf{Type 5: Under-Segmented Speech Balloon Annotations}, where multiple connected speech balloons are annotated as a single text region.
    Images courtesy of Yamada Uduki.}
    \label{fig:type5}
\end{figure*}

\paragraph{Type 5: Under-Segmented Speech Balloon Annotations}
This type refers to cases where multiple connected speech balloons are annotated using a single bounding box (see Figure~\ref{fig:type5}).
While such annotations were acceptable under earlier annotation practices, they can become problematic under modern OCR evaluation settings. 
In particular, recent OCR systems tend to detect such regions as separate text instances.
This mismatch may cause correct detections to be incorrectly penalized during evaluation.

To address this issue, we identify cases where a single annotation bounding box overlaps with multiple OCR-detected bounding boxes. 
These cases are manually verified, and the original bounding box is divided into multiple regions corresponding to individual speech balloons.

This revision enables more accurate evaluation of OCR systems by better aligning annotations with how modern AI systems interpret manga text regions and speech balloon structures.

\begin{table}[t]
    \centering
    \caption{OCR evaluation results before and after annotation revision. Following MangaOCR~\cite{baek2025mangavqa}, we report end-to-end (E2E) precision, recall, and H-mean scores.}
    \label{tab:ocr_eval}
    \begin{tabular}{lccc}
        \toprule
        \textbf{Annotations} & \textbf{Precision} & \textbf{Recall} & \textbf{H-mean} \\
        \midrule
        Original Manga109 & 46.5 & 50.6 & 48.5 \\
        Manga109-v2026 & 63.4 & 62.4 & 62.9 \\
        \bottomrule
    \end{tabular}
\end{table}

\subsection{Impact of Annotation Revisions on OCR Evaluation}
To examine the effect of our revisions, we evaluate the OCR outputs described in Section~\ref{sec:issue_detection} using both the original Manga109 annotations and the revised Manga109-v2026 annotations. 
Following the evaluation protocol used in MangaOCR~\cite{baek2025mangavqa}, we evaluate the same OCR outputs using both the original Manga109 annotations and the revised Manga109-v2026 annotations. 
As shown in Table~\ref{tab:ocr_eval}, the revised annotations substantially improve OCR evaluation performance, increasing the E2E H-mean score from 48.5 to 62.9 (+14.4 percentage points).
This improvement suggests that the revised annotations better align with how modern OCR systems interpret manga text regions and speech balloon structures.

These results suggest that revisiting legacy annotations can improve the reliability of OCR evaluation in modern manga AI settings. 
More broadly, they demonstrate the importance of updating culturally grounded datasets to better reflect how contemporary AI systems process expressive manga text and layouts.

\section{Conclusion}
In this work, we revisited the dialogue text annotations of Manga109 and identified five categories of annotation issues affecting modern manga AI tasks.
Through OCR-assisted issue detection and manual revision, we constructed Manga109-v2026 with approximately 29,000 revised dialogue annotations.
Experimental results showed that the revised annotations substantially improve OCR evaluation performance, demonstrating that annotation practices acceptable under earlier settings may no longer align well with modern AI-based manga analysis. 
Our findings highlight the importance of continuously revisiting culturally grounded datasets as AI systems and downstream tasks evolve.
We hope Manga109-v2026 serves as a more reliable foundation for future research on manga OCR, translation, and multimodal manga understanding.